\documentclass[letterpaper, 10 pt, conference]{ieeeconf}  

\usepackage[utf8]{inputenc}

\IEEEoverridecommandlockouts

\usepackage{amsmath,amssymb,amsfonts}
\usepackage{algorithmic}
\usepackage{graphicx}
\usepackage{textcomp}
\usepackage{array}

\usepackage{graphicx}
\usepackage{float} 
\usepackage{blindtext}
\usepackage{afterpage}
\usepackage{hhline,caption}
\usepackage{hyperref}
\usepackage{comment}

\usepackage[utf8]{inputenc}
\usepackage{subcaption}
\usepackage{xcolor}

\usepackage[ruled]{algorithm2e} 
\usepackage{cite}
    
\author{Wen Fan*, Haoran Li*, Weiyong Si, Shan Luo, Nathan Lepora, Dandan Zhang*
}
\begin{document}

\title{\LARGE \bf
ViTacTip: Design and Verification of a Novel Biomimetic Physical Vision-Tactile Fusion Sensor

\thanks{W. Fan, D. Zhang are with the Department of Bioengineering, Imperial College London. H. Li, W. Si, N. Lepora are with the Department of Engineering Mathematics, University of Bristol. S. Luo is with the Department of Engineering, King's College London. Corresponding: Dandan Zhang, d.zhang17@imperial.ac.uk}
}

\maketitle

\begin{abstract}
Tactile sensing is significant for robotics since it can obtain physical contact information during manipulation. To capture multimodal contact information within a compact framework, we designed a novel sensor called ViTacTip, which seamlessly integrates both tactile and visual perception capabilities into a single, integrated sensor unit. ViTacTip features a transparent skin to capture fine features of objects during contact, which can be known as the see-through-skin mechanism. In the meantime, the biomimetic tips embedded in ViTacTip can amplify touch motions during tactile perception. For comparative analysis, we also fabricated a ViTac sensor devoid of biomimetic tips, as well as a TacTip sensor with opaque skin. 
Furthermore, we develop a Generative Adversarial Network (GAN)-based approach for modality switching between different perception modes, effectively alternating the emphasis between vision and tactile perception modes. We conducted a performance evaluation of the proposed sensor across three distinct tasks: i) grating identification, ii) pose regression, and iii) contact localization and force estimation.
In the grating identification task, ViTacTip demonstrated an accuracy of 99.72\%, surpassing TacTip, which achieved 94.60\%. It also exhibited superior performance in both pose and force estimation tasks with the minimum error of 0.08~mm and 0.03N, respectively, in contrast to ViTac's 0.12~mm and 0.15N. Results indicate that ViTacTip outperforms single-modality sensors.
\end{abstract}

\section{Introduction}

In real-world scenarios, a single sensing modality or data acquisition method may prove inadequate for achieving a comprehensive understanding of the complex environment. Thus, there has been a notable surge in research aimed at integrating visual and tactile sensing through representation learning. These approaches encompass many strategies, such as combining features acquired from visual and tactile sensors within a shared latent space to augment robot perception \cite{luo2018vitac} \cite{yuan2017connecting}. Alternatively, some methods focus on mapping representations from one sensory modality to another for cross-modal perception~\cite{lee2019touching} or leverage information from one modality to guide exploration in another \cite{jiang2022shall}. In these studies, visual cameras are employed for capturing visual data, while tactile sensors are utilized to gather contact information. Such integrative approaches can pave the way for more robotic systems.

\begin{figure}[t]
	\centering
\captionsetup{font=footnotesize,labelsep=period}
\includegraphics[width = 0.9\hsize]{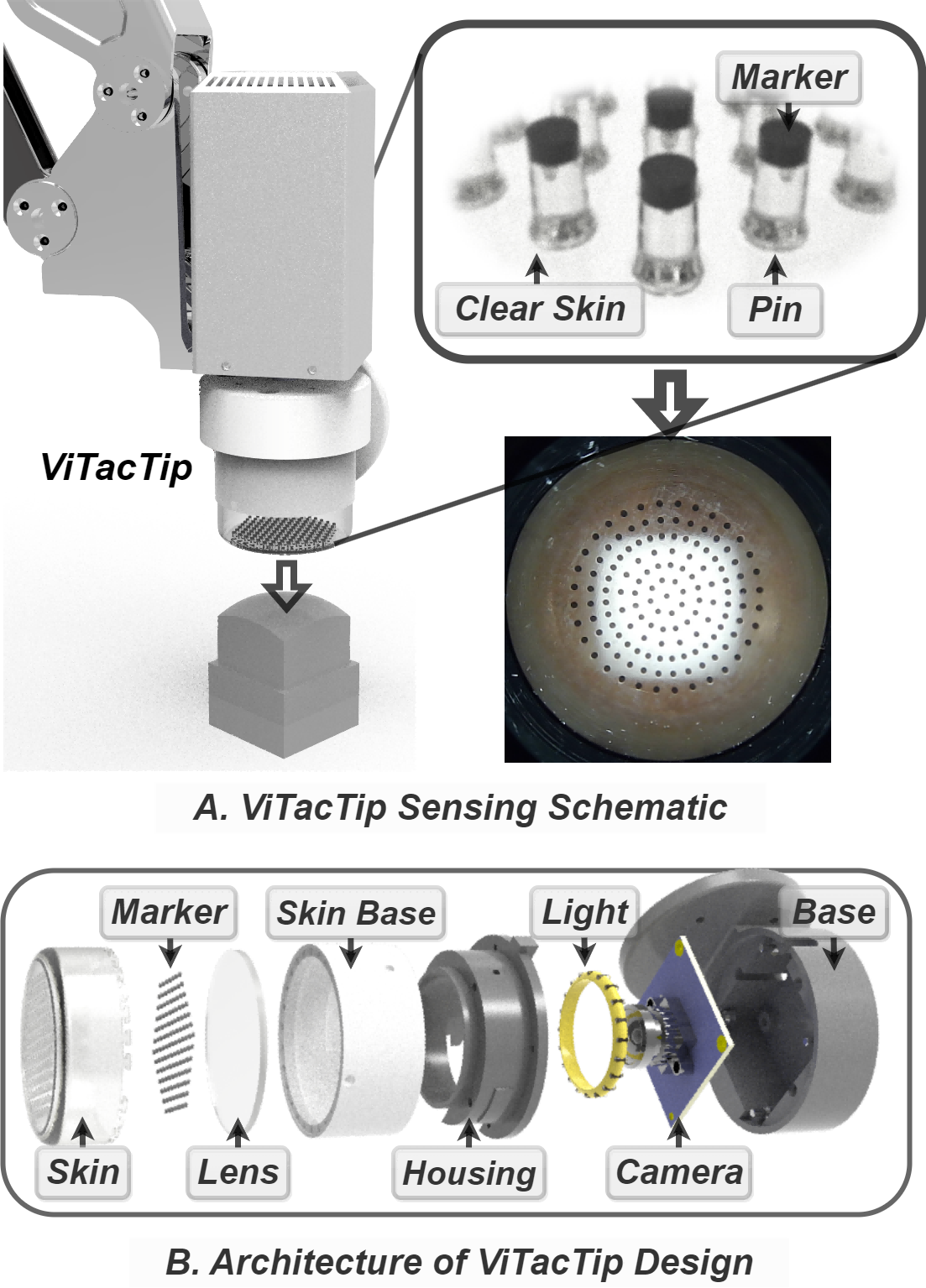}
	\caption{A: ViTacTip schematic which achieves an internal fusion of vision and tactile modalities through the structure of bio-inspired pins and transparent skin. B: Architecture of ViTacTip sensor, which illustrates exploded view of sub-parts.}
 \vspace{-0.6cm}
	\label{ViTacTip Principle}
\end{figure}

Traditional methods in robotics, exemplified by the `eye-in-hand' approach \cite{jang1991feature}, typically involve equipping the robot's end-effector with a camera to gather close-view visual data and contact information. Despite its widespread use, this approach is susceptible to occlusion and incurs additional computational load due to the need for switching between different sensing modalities. These issues may affect its effectiveness in applications demanding precise grasping and dexterous manipulation \cite{9813397}.
In contrast, employing a single sensor capable of capturing both visual and tactile information presents substantial advantages. This approach helps minimize the complexity of the hardware setup and reduces computational demands in robotic systems.

Recently, Vision-Based Tactile Sensors (VBTSs) have been developed \cite{yuan2017gelsight}, which utilize an internal camera to observe the deformation of a soft elastomer when contact occurs. VBTSs offer higher spatial resolution compared to tactile sensors based on other sensing principles. Typically, VBTSs have three representative types: (1) marker-based systems like TacTip \cite{ward2018tactip}, (2) sensors equipped with reflective coatings designed to capture comprehensive deformation information, such as GelSight\cite{yuan2017gelsight}, and (3) vision-tactile fusion sensors utilizing marker-embedded silicone, such as FingerVision\cite{yamaguchi2016combining}. Although VBTSs use cameras to obtain tactile information, their design, such as the black skin of TacTip\cite{ward2018tactip} or the reflective layer of GelSight\cite{yuan2017gelsight}, prevents them from concurrently gathering visual data about their surroundings. In contrast, FingerVision\cite{yamaguchi2016combining} benefits from rich visual information, providing additional perceptual modality, yet is less resistant and robust to interference, such as external light.

In this study, we introduce a novel VBTS sensor named ViTacTip, which integrates both vision and tactile perception capabilities into a single unit. A Generative Adversarial Network (GAN)-based approach is designed to switch seamlessly between capturing visual information from the environment and gathering tactile data upon contact with the targeted object. The key contributions are as follows:

i) The innovative development of the ViTacTip sensor, a vision-tactile fusion device. This sensor stands out for its dual-attribute data acquisition capability. It is adept not only at collecting tactile information but also at capturing visual features such as the color and geometry of objects it contacts.

ii) The implementation of a Generative Adversarial Network (GAN)-based methodology to enhance modality switching between visual and tactile sensing. This approach proves particularly effective in mitigating the challenges posed by variable ambient lighting conditions and in improving the visualization of contact interactions. 

\section{Related Work}

Reflective-based sensors, such as GelSight \cite{yuan2017gelsight}, and marker-type sensors, such as TacTip \cite{ward2018tactip}, are the most popular VBTSs among existing literature.
The GelSight sensor features a coated, flat, and relatively rigid sensing skin, whereas the TacTip sensor has a soft, flexible, dome-like sensing skin. Sensors similar to GelSight possess flat sensing surfaces made of molded elastomer, which are side-lit and excel in capturing detailed textures of contacted objects. Currently, Gelsight-type VBTSs with non-flat structures are also designed, such as GelTip \cite{gomes2020geltip}, OmniTact \cite{padmanabha2020omnitact}, and GelSight360 \cite{tippur2023gelsight360}. Nonetheless, GelSight and TacTip are limited in their ability to incorporate a vision modality, as the opaque skin and reflective coating block the environmental light. 
In contrast, FingerVision \cite{yamaguchi2016combining}, which utilizes optically transparent silicone, offers the distinct advantage of directly acquiring visual information through its `skin'. However, in FingerVision, both visual and tactile information are closely intertwined. This close coupling can potentially lead to mutual interference between the two modalities. As a consequence, FingerVision's perception capability may be compromised, particularly under varying environmental lighting conditions. 


Given the intricacies involved in the fusion of separate visual and tactile data, some preliminary work demonstrated the physical integration of visual and tactile information. Inspired by the properties of unidirectional transparent glass, some researchers applied a transparent, thin coating to the elastomer surface \cite{hogan2021seeing, athar2023vistac, roberge2023stereotac}.  This approach modulates sensing modalities by adjusting the internal lighting.
Specifically, when internal lights are deactivated, external light becomes predominant, enabling the embedded camera to capture visual data from the environment. In contrast, with the internal lights turned on, the camera shifts its focus to exclusively gather tactile information, as the enhanced internal illumination negates the camera's ability to perceive external visual cues. This method effectively balances the dual requirements of capturing external visual data and tactile information, depending on the lighting conditions.

Other sensors capable of perceiving both visual and tactile information, such as SpecTac \cite{wang2022spectac} and Tac \cite{shimonomura2016robotic}, depend heavily on well-lit environments for effective modality conversion. Their performance notably declines in low-light conditions. In contrast, TIRgel  \cite{zhang2023tirgel} demonstrates reliable sensing performance even in dark environments. This sensor combines a transparent elastomer with a focus-adjustable camera, which can enable the switch between visual and tactile modalities by employing total internal reflection (TIR) within the elastomer for tactile representation.
However, TIRgel shares similar material properties and design elements with GelSight-type sensors, primarily featuring a relatively flat surface. This feature and the TIR mechanism requirement may limit its flexibility of customization like TacTip-type sensors, which are distinguished by a pliable and curved sensing surface \cite{lepora2022digitac}. Consequently, TIRgel may face limitations in dynamic manipulation.

In summary, our research aims to develop a multi-modal sensor that integrates vision and tactile capabilities. ViTacTip is envisioned to combine the strengths of existing technologies while overcoming their limitations, enabling more effective and adaptable performance in a wider range of applications, particularly in dynamic manipulation tasks.


\section{Design and Fabrication}

\subsection{ViTacTip Design Principles}

Inspired by the structural design of the TacTip sensor \cite{ward2018tactip}, the ViTacTip sensor is designed to replicate the layered macro-structure typical of human fingertip skin, illustrated in Fig. \ref{ViTacTip Principle}(A). The key components of the sensor consist of a thin, flexible, and transparent rubber-like skin. This skin is designed to feature structural details reminiscent of those in the glabrous epidermis. This design choice endows the sensor with exceptional versatility and sensitivity when interacting with various environments.  
The tactile skin of the ViTacTip sensor is fabricated using Agilus30 Clear, a material selected for its notable flexibility and high light transmission properties. 
An internal camera is positioned to track pin-like markers on the rubber skin. This design enables the sensor to perform highly sensitive detection of surface deflections, which can ensure capturing a wide range of tactile interactions with remarkable accuracy.

\subsection{ViTacTip Fabrication}
Skin: To fabricate ViTacTip's skin, we adopted a multi-material 3D printing approach. This method eliminates the need for the time-consuming procedure of mould-making, silicone rubber skin coating, and paint application, thereby leading to better cost-efficiency and design reproducibility.  Moreover, we integrated gel-like materials into the transparent outer layer and the lens of the sensor. This gel is designed to replicate human skin's elasticity, enhancing the sensor's ability to provide realistic feedback. Biomimetic tips are printed on the skin as markers \cite{winstone2012tactip}.
This approach is taken to ensure that ViTacTip's skin not only performs effectively in sensing tasks but also mimics the natural properties of human skin, which is vital for sensitive tactile responses.

Illumination: Uniform illumination across the tactile skin's surface is essential to accurately capture deformations. Direct vertical illumination is not suitable because it could interfere with the camera's functionality. Similarly, lateral illumination is problematic as it could cause uneven brightness at the edges of the sensor. To address these challenges, we incorporated a flexible, ring-shaped light source around the camera. 
This design, shown in Fig.~\ref{ViTacTip Principle} (B), ensures consistent lighting across the sensor's surface. It also allows for easy adjustment of the distance between the lens and the light source, which is beneficial in confined spaces within the tactile sensor.

Camera: The choice of camera for ViTacTip is critical due to spatial constraints. A camera with a broad field of view (FoV) and a short focal length is needed. A wide FoV ensures comprehensive coverage of the observation area, while a short focal length allows for a compact sensor design by reducing the distance needed between the tactile skin and the camera. For these reasons, we selected the ELP camera (USBFHD06H-L180), which is equipped with a wide-angle lens, making it ideal for our sensor's requirements \cite{ward2018tactip}.

\subsection{Vision-Tactile Fusion Imaging Principle}

The proposed ViTacTip offers a distinctive advantage in vision-tactile fusion by representing tactile information through visual features.   The ring-shaped light source emits uniform light, ensuring the visibility of black markers located at the tips of pins within the camera's field of view. Similar to the TacTip sensor, when ViTacTip contacts an object, like an indenter, its soft skin deforms to match the object's surface. This deformation, supported by the internal gel, changes the direction of the pins, thus creating a tactile representation.

In addition to tactile sensing, ViTacTip’s transparent skin allows a portion of the light to illuminate an area of about 20~mm beneath the sensor. This feature enables the sensor to capture visual information about objects in this region, including their outline, color, and position. When the sensor is within 5-10 mm of an object, it can identify most of its features.  Upon direct contact, ViTacTip is capable of detailed texture analysis.  The reflected light, bearing both visual and tactile data, passes through the skin and is recorded by the camera. This process results in a comprehensive vision-tactile fusion image, offering a more detailed perception than tactile data alone.

\begin{figure}[!htbp]
	\centering
\captionsetup{font=footnotesize,labelsep=period}
	\includegraphics[width = 0.9\hsize]{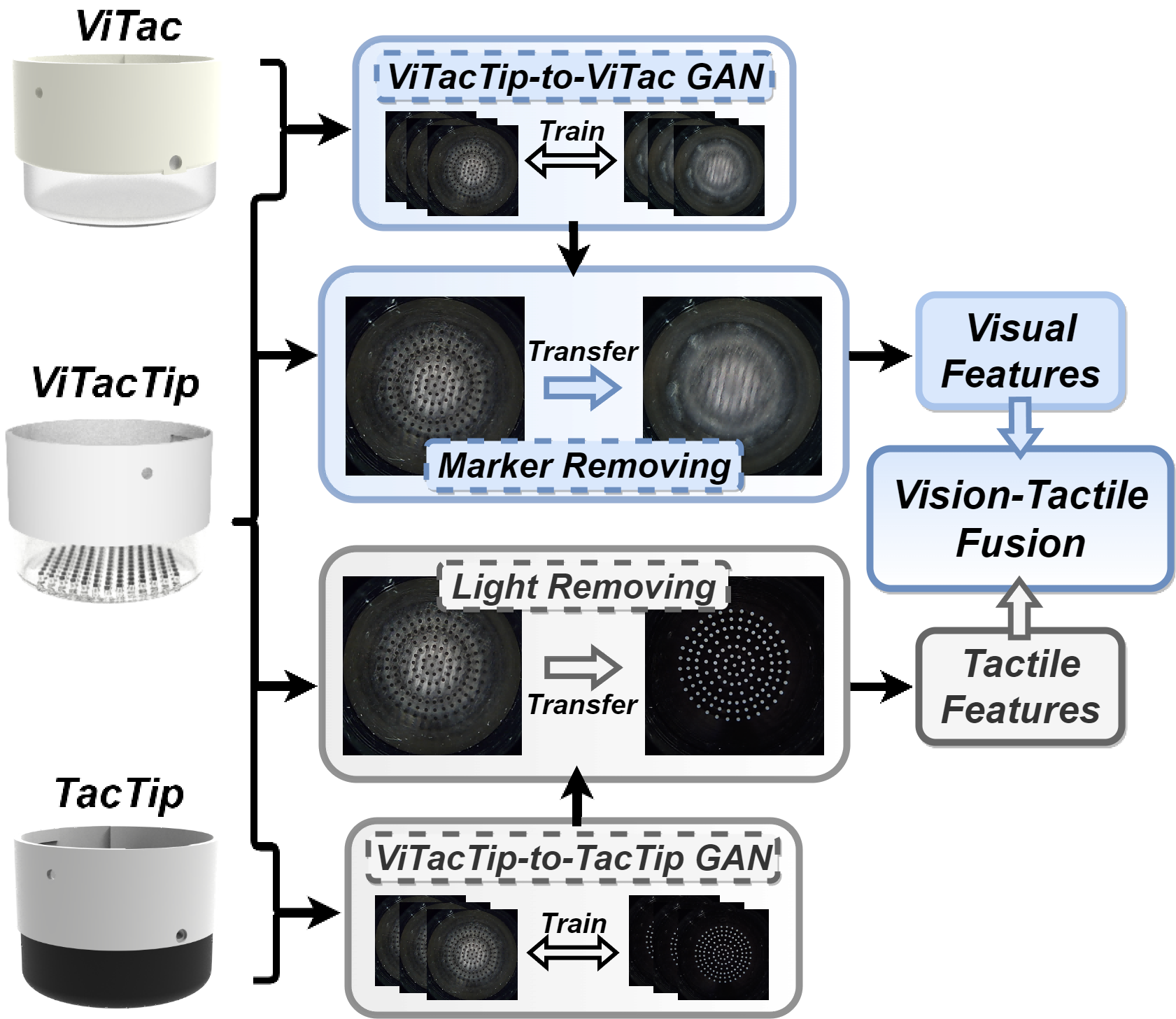}
	\caption{Modality conversion framework. Two independent GAN models are trained with three datasets from ViTac, ViTacTip, and TacTip separately to achieve marker-removing tasks and light-removing tasks.}
  \vspace{-0.4cm}
	\label{gan transfer}
\end{figure}

\begin{figure*}[!htbp]
\captionsetup{font=footnotesize,labelsep=period}
	\centering
\includegraphics[width = 0.95\hsize]{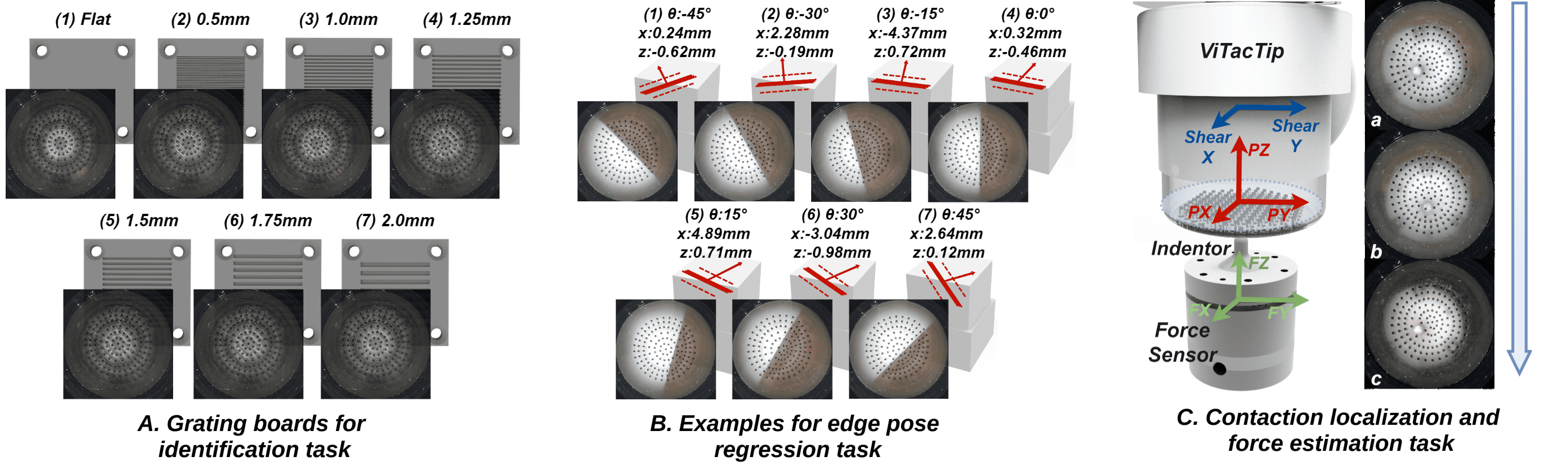}
	\caption{Overview of three experiments designed for ViTacTip, ViTac, and TacTip. }
	\label{task_setup}
\end{figure*}

\begin{figure*}[!htbp]
\captionsetup{font=footnotesize,labelsep=period}
	\centering
	\includegraphics[width = 1\hsize]{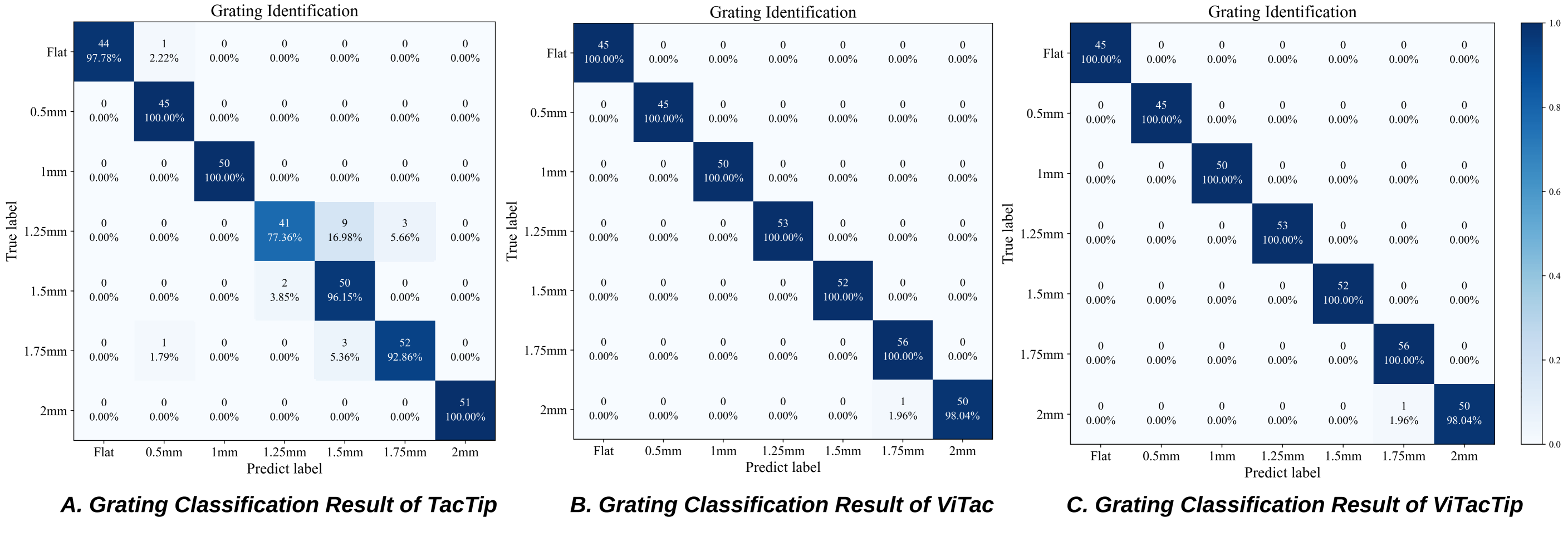}
	\caption{Evaluation result of TacTip, ViTac, and ViTacTip on grating identification.}
 \vspace{-0.3cm}
	\label{grating evaluation}
\end{figure*}

\subsection{Modality Conversion Principle}

The physical multi-modality fusion approach used in ViTacTip has both advantages and disadvantages. For example,  while the sensor's marker layout may not be sufficiently dense to capture fine tactile textures, the integration of the visual modality offsets this by providing additional details like color. However, the inclusion of black markers can introduce visual noise, potentially distorting visual perception. 
 Additionally, despite having an internal light source, the sensor's transparent skin is susceptible to external lighting interference. The colorful background complicates conventional tactile processing techniques such as marker detection and tracking, which presents a challenge in data interpretation.

To address the challenges identified with the ViTacTip sensor, we implement a Pix2Pix Generative Adversarial Network (GAN) approach \cite{isola2017image} and construct two unique GANs for different purposes. One network is dedicated to marker removal, while the other focuses on correcting for ambient light interference. Adhering to the hyperparameters outlined in the original Pix2Pix study \cite{isola2017image}, our framework's schematic is illustrated in Fig.~\ref{gan transfer}. For ground truth data generation and comparative analysis, two additional tactile sensors were employed.
The first sensor, the \textbf{TacTip} \cite{lepora2021soft}, features pin-shaped markers embedded in opaque skin. This feature effectively protects it from external light sources. The second sensor, named \textbf{ViTac}, is equipped solely with transparent skin. Unlike the ViTacTip, it lacks internal pins and markers, providing a distinct design for comparative evaluation.

In our experiment, we gathered three distinct datasets, each utilizing one of the tactile sensors independently. The first unique GAN was trained to transform data from ViTacTip into a format akin to that of ViTac. This transformation effectively removes any occlusions caused by markers in the ViTacTip data, creating outputs similar to ViTac's. Therefore, we term this the ViTacTip-to-ViTac GAN. The second GAN model is tailored to convert data from ViTacTip to resemble that of TacTip. This conversion process effectively extracts markers and mitigates the effects of ambient light interference.
 We call this the ViTacTip-to-TacTip GAN. 
 
 The essence of this approach is the distinct extraction of visual and tactile data, which were initially integrated into ViTacTip. By disentangling these two features, we enhance the sensor's versatility and adaptability for diverse applications, overcoming the inherent limitations of the original fused data format.

\begin{figure*}[!htbp]
\captionsetup{font=footnotesize,labelsep=period}
	\centering
	\includegraphics[width = 0.82\hsize]{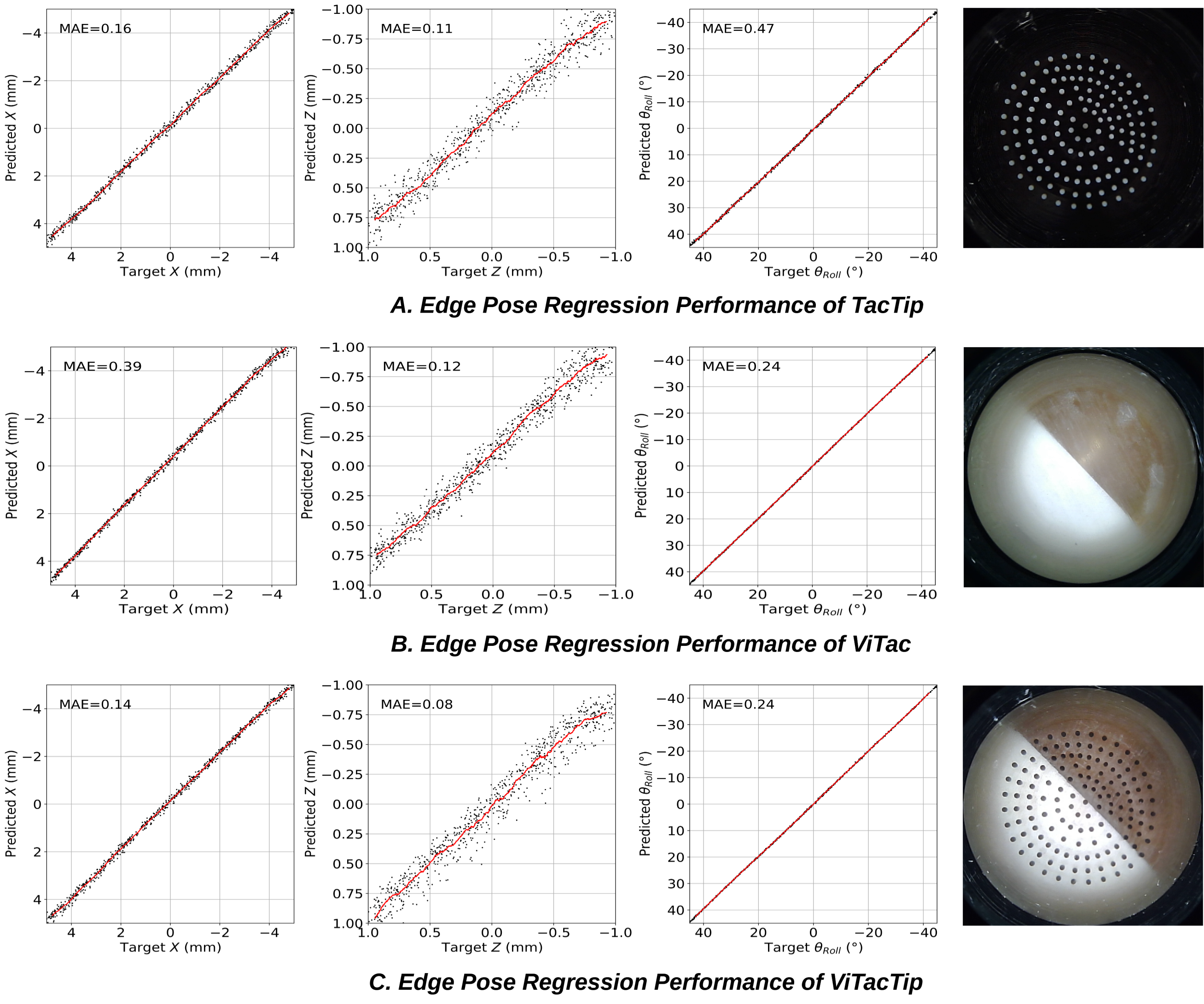}
	\caption{Evaluation result of TacTip, ViTac, and ViTacTip on edge pose regression for horizontal distance $X$, press depth $Z$, rotation angle $\theta$. The red line indicates the mean fit of regression values, the smaller the deviation of this line from the diagonal $y=x$, the better the prediction is proved to be. }
 \vspace{-0.4cm}
	\label{pose error}
\end{figure*}

\section{Experiment and Results}
In our study, we conducted a series of experiments to evaluate the performance of the ViTacTip sensor in various tasks. These tasks included grating identification, edge pose regression, force estimation, and contact localization. TacTip and ViTac sensors served as baselines for comparison.
The experimental data for these tasks were collected using a desktop robot arm (MG400, Dobot), employing a methodology similar to that described in \cite{lepora2022digitac}. For data processing, we adopted a densely connected convolutional network, DenseNet121 \cite{huang2017densely}. This network was chosen for its proven efficiency in extracting critical information from images, making it particularly suitable for analyzing the complex data sets generated in our experiments.

\subsection{Grating Identification}
To determine the spatial resolution capabilities of the ViTacTip sensor, our study incorporated an approach using objects with varying levels of detail, inspired by the experimental setup in Yuan et al. \cite{yuan2017gelsight}. More specifically, we employed a series of grating boards, each characterized by different line densities. As depicted in Fig.~\ref{task_setup} (A), we classified the spatial resolution of these boards into three distinct groups for tactile sensing assessment:
i) Millimeter group: with line spacings of 0 mm, 1 mm, and 2 mm;
ii) Half-millimeter group: featuring finer line spacings of 0 mm, 0.5 mm, 1 mm, 1.5 mm, and 2 mm;
iii) Quarter-millimeter group: comprising the most detailed line spacings of 1 mm, 1.25 mm, 1.5 mm, 1.75 mm, and 2 mm.

We gathered a substantial dataset consisting of 3500 data points, with 500 data points for each of the seven categories of line spacings. The data were divided into a 7:2:1 ratio for training, validation, and testing, respectively. This approach allowed for a rigorous evaluation of the ViTacTip sensor's sensitivity and accuracy in discerning fine spatial details across a range of resolutions.
The results of this evaluation are presented in Fig. \ref{grating evaluation}. According to Fig. \ref{grating evaluation}(A), TacTip struggled with the quarter-millimeter measurements, particularly with identifying 1.25 mm, 1.5 mm, and 1.75 mm, resulting in a test accuracy of 94.60\%. In contrast, both ViTac and ViTacTip were more successful, achieving a high accuracy of 99.72\%, as depicted in Fig.~\ref{grating evaluation} (B) and (C). The improvement in spatial resolution for ViTacTip, when compared to TacTip, is primarily due to the use of visual features that help overcome the limitations of TacTip associated with marker density and skin conformability.

\subsection{Pose Regression}

To effectively evaluate a tactile sensor's proprioception capabilities,  several necessary tests should be conducted to assess the sensor's feedback on contact position and orientation interacting within its environment, as recommended by \cite{lepora2021pose}. Specifically, our study included an edge pose regression test which involved estimating the sensor's pose relative to a square stimulus's boundary.  The parameters measured were the horizontal distance $X$ from the sensor's center to the boundary, the depth of press $Z$, and the angle of rotation $\theta$ along the Z-axis, as illustrated in Fig.~\ref{task_setup} (B).

We collected 3000 images for this purpose, ensuring the edge pose values ($X$, $Z$, $\theta$) varied within the ranges of [-5,5]~mm, [-1,1]~mm, and [-45,45]~degrees, respectively. We allocated 75\% of these images for training and the rest for validation and testing. The comparative performance of three sensors (TacTip, ViTac, and ViTacTip) is detailed in Fig.~\ref{pose error}.

The results show that TacTip and ViTacTip, equipped with pins and markers, are more effective in amplifying deformations caused by skin pressing against the edge. This feature enabled them to outperform ViTac in predicting the horizontal $X$ and depth $Z$. Furthermore, ViTac and ViTacTip are similarly effective and surpassed TacTip in determining the rotation angle $\theta$. This advantage is attributed to their visual features, which allow for the visual determination of the edge's orientation, as clearly shown in Fig.~\ref{task_setup} (B)/(C).

\begin{figure*}[!htbp]
	\centering
\captionsetup{font=footnotesize,labelsep=period}
\captionsetup{font=footnotesize,labelsep=period}
	\includegraphics[width = 1\hsize]{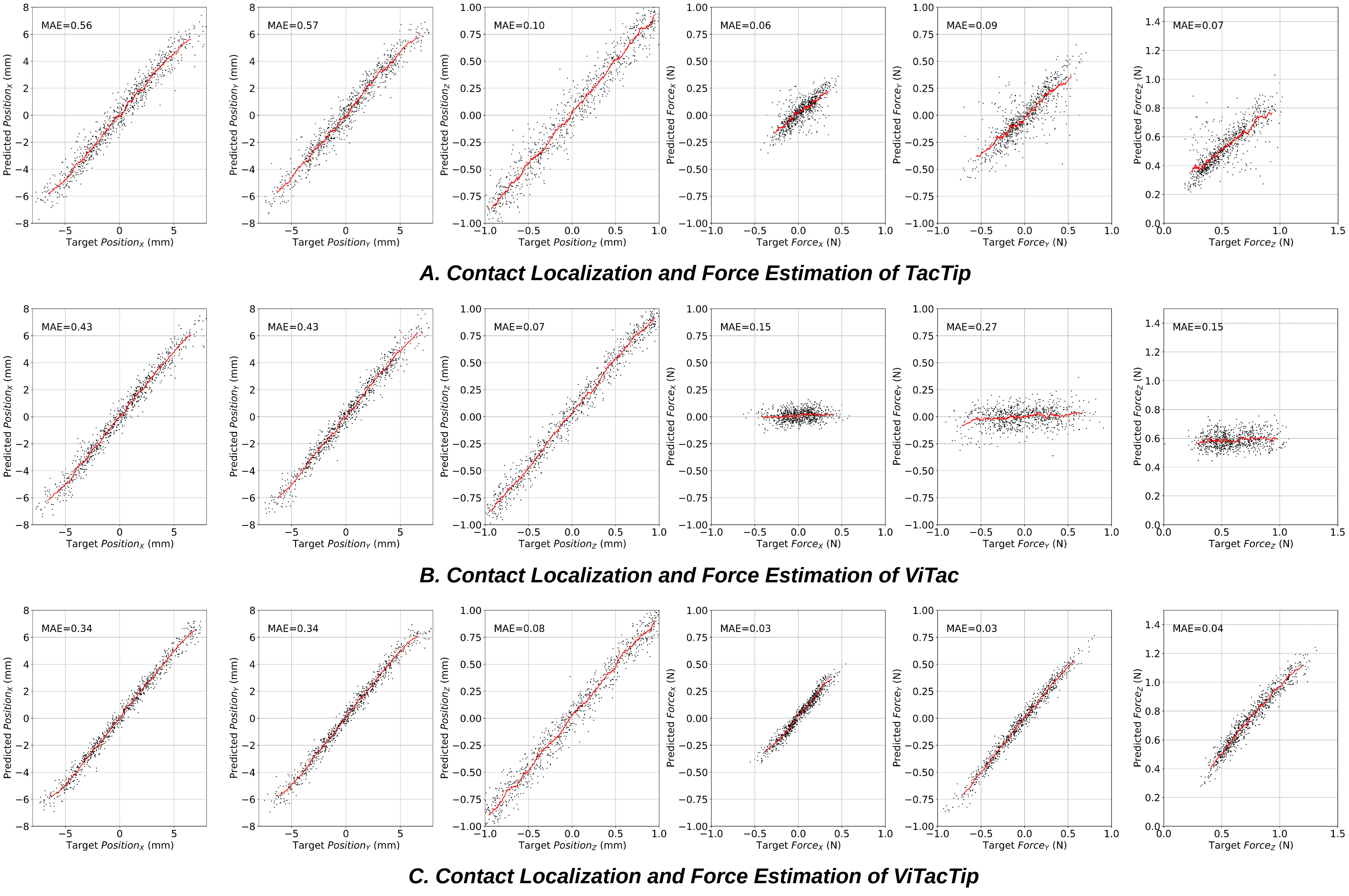}
	\caption{Performance of TacTip, ViTac, ViTacTip on contact localization ($Px$, $Py$, $Pz$) and force estimation ($Fx$, $Fy$, $Fz$). The red line indicates the mean fit of regression values, the smaller the deviation of this line from the diagonal $y=x$, the better the prediction is proved to be estimated.}
 \vspace{-0.3cm}
	\label{force}
\end{figure*}

\subsection{Contact Localization and Force Estimation}
Force sensors play an important role in robotics for practical real-world applications. However, these sensors are often expensive.  For example, standard force sensors such as the ATI (Axia80-M20) or NBIT typically range in price from \$3,000 to \$5,000, making them a significant investment. Consequently, there is an increasing interest in exploring tactile sensors as a more cost-effective substitute for traditional force sensors.

In our research, we designed an experiment to evaluate the ViTacTip sensor's capability to function as a substitute for conventional force sensors. The experimental setup, depicted in Fig.~\ref{task_setup} (C), was structured similarly to the pose regression test we conducted earlier. However, a modification was made: the lower square stimulus used in the pose regression test was replaced with an NBIT force sensor. This adaptation allowed us to directly compare the force measurements obtained from the ViTacTip sensor with those from a standard force sensor.  


During the data collection phase, we generate a comprehensive force profile on the ViTacTip sensor. After an initial press on the sensor, we introduced shear displacement along the horizontal plane (X/Y), thereby inducing a three-dimensional force upon the sensor's surface. This methodology was designed to simulate real-world conditions and assess the stability of the sensor's performance under varying circumstances.
To further mimic realistic environments and rigorously evaluate the sensor's adaptability, we varied the intensity of external light sources, thereby altering the ambient light conditions during the experiment. This variation was crucial to assess the robustness of force estimation in real-world scenarios.
We aimed to measure the forces (denoted as $Fx$, $Fy$, $Fz$) generated by both the normal pressing and the horizontal shearing actions. In the meantime, we predict the indentor's relative spatial location ($Px$, $Py$, $Pz$) at the contact point with the sensor skin.

The results, shown in Fig. \ref{force}, indicate that ViTacTip can precisely localize contact points, even under varying shear conditions and lighting changes. Furthermore, it successfully predicted three-dimensional forces, with a horizontal force estimation error within a range of 0.03~N (between -0.5 and 0.5~N) and a normal pressure prediction error of 0.04~N (between 0.4 and 1.2~N).
 Compared to TacTip, ViTacTip showed superior performance in contact localization. This improvement can be attributed to the additional visual information provided by ViTacTip, which reduces positioning errors after shear movements. In contrast, ViTac's lack of markers leads to less accurate force mapping. These findings highlight the benefits of integrating visual and tactile sensing in ViTacTip.

\section{Conclusions and Future Work}
\label{Conclusions}
 
In our study, we introduce ViTacTip, a novel sensor with multi-modality capabilities.  In performance comparisons with TacTip, ViTacTip shows improvements on reducing pose regression error by 46\%, contact localization error by 60\%, force estimation error by 52\%, and increasing grating classification accuracy by 5\%.
Furthermore, our study incorporates the use of two specialized GANs, which are designed to facilitate a modality switch that allows the ViTacTip sensor to emulate the sensing characteristics of both TacTip and ViTac. 

In future work, we plan to rigorously test the ViTacTip sensor under varied environmental conditions to validate its robustness and adaptability. We will integrate ViTacTip into robotic hands for executing dexterous manipulation tasks that require advanced multi-modality perception.
ViTacTip holds potential for enhancements and adaptations for a broad range of applications beyond its current demonstrations. 

\bibliographystyle{IEEEtran}

\bibliography{IEEEabrv,ref}

\end{document}